\documentclass{article}
\usepackage{acra}
\usepackage{graphicx}
\usepackage{multirow}
\usepackage{adjustbox}
\usepackage{amsmath}
\usepackage{url}
\usepackage{hyperref}
\usepackage{array}

\graphicspath{{images/}}

\title{Pallet Detection from Synthetic Data using Game Engines}
\author{Jouveer Naidoo$^{*}$, Nicholas Bates$^{*}$,\\ \textbf{Trevor Gee, Mahla Nejati$^{**}$} \\
Centre for Automation and Robotic Engineering Science\\
The University of Auckland, New Zealand\\
$^{*}$\{jnai894, nbat773\}@aucklanduni.ac.nz, $^{**}$m.nejati@auckland.ac.nz}

\begin{document}

\maketitle

\begin{abstract}

This research sets out to assess the viability of using game engines to generate synthetic training data for machine learning in the context of pallet segmentation. Using synthetic data has been proven in prior research to be a viable means of training neural networks, and saves hours of manual labour due to the reduced need for manual image annotation. Machine vision for pallet detection can benefit from synthetic data as the industry increases the development of autonomous warehousing technologies. As per our methodology, we developed a tool capable of automatically generating large amounts of annotated training data from 3D models at pixel-perfect accuracy and a much faster rate than manual approaches. Regarding image segmentation, a Mask R-CNN pipeline was used, which achieved an AP50 of 86\% for individual pallets.

\end{abstract}

\section{Introduction}

Supervised machine learning typically requires a large amount of training data that is often manually labelled. Manual labelling is tedious, work-intensive, expensive, and error-prone. To illustrate this, in the context of the pallet segmentation problem that is the focus of this research, a small set of 70 images of pallets in a warehouse took one person approximately three hours to complete. 

Of course, if synthetic data resulted in models of equivalent quality to ones generated from manually labelled data, this could overcome many problems associated with manually labelled data. Additionally, this notion is greatly encouraged because modern graphic engines can now generate high-quality ``photorealistic'' renderings.

The aims of this research are to validate the feasibility of using synthetically generated pallets in order to train a neural network to operate on real images. This is useful for our research partner who is developing a self-driving forklift system where the localisation of pallets in images would aid in the movement planning of forklifts, as well as other autonomous vehicles in general.

A State-of-the-art graphics engine will be used to generate realistic models of warehouses and pallets, along with accurate labels and annotations, to train a neural network capable of detecting and classifying pallets in real-world scenarios. In addition, it identifies certain factors that affect the performance of detection and classification of pallets when using synthetic data.

The main driver for this work was the pallet detection problem, which is an essential step toward fully autonomous factory floors.

\section{Related Work}
This part of the paper focuses on existing research into machine vision, synthetic data and pallet detection. Existing literature broadly falls into these three categories and is crucial in order to define a starting point for the project.

\subsection{Machine Vision}
This section presents an overview of current and recent deep learning object detection methodologies, models, and tools in order to establish a background on how pallets are to be detected with machine learning.

Many object detection problems involve identifying and classifying several common objects, such as people and vehicles \cite{Lin2014}. In order to train models to do this, image data-sets such as MS COCO \cite{Lin2014} and Open Image \cite{Kuznetsova2020} exist, which together contain millions of images and annotations for everyday objects \cite{Zaidi2022}. It is thus possible to use these sets to train a neural network for simple object detection \cite{Nejati2019a,Williams2019}.

A number of different object detectors are used, such as YOLOv4 \cite{Bochkovskiy2020} from the YOLO (You Only Look Once) \cite{Redmon2016} family. Swin-L \cite{Liu2021} was found by \cite{Zaidi2022} to have the highest average precision of 57.70\% on MS COCO at the time of publication. “While CNNs have been the backbone on advancement in vision, they have some inherent shortcomings” \cite{Zaidi2022}, which is what Swin-L overcomes by the use of transformers \cite{Vaswani2017}.

Some more popular models, which are supported by tools such as Detectron2 \cite{Wu2019} and MMDetection \cite{mmdetection}, include R-CNN \cite{Girshick2014}, Fast R-CNN \cite{Girshick2015}, Faster R-CNN \cite{Ren2017} and Mask R-CNN \cite{He2020}, which are all in the Region-Convolutional Neural Network (R-CNN) family, as well as Single Shot MultiBox Detector \cite{Liu2016}. These models could be looked into more for future work with the paper, as certain models may offer better detection rates than others.

Machine vision is currently solved using machine learning, which requires training data in all cases.

\subsection{Pallet Detection}
One of the most common methods of detecting pallets was through the use of a monocular vision sensor, also known as a camera \cite{Zaccaria2020,Li2019,Syu2017,Seelinger2005,Pages2001,Garibotto1996,Cui2010,Chen2012}. A neural network is then used, typically with R-CNN \cite{Zaccaria2020}, YOLOv4 \cite{Zaccaria2020} or SSD \cite{Zaccaria2020,Li2019}. In\ \cite{Zaccaria2020}, it was found that R-CNN and SSD perform better than YOLOv4.

\cite{Zaccaria2020} found that they achieved a successful detection rate using a three-stage method of running an image through the CNN, which would detect pallet pockets and faces before finally linking the pockets to faces. They could even take into account pallets wrapped in plastic film, as well as varying elevations, racks, and orientations. This was successful with a camera resolution of 3280x2464, with 1344 images: 991 training and 353 testing \cite{Zaccaria2020}.

Another common method of pallet detection uses laser scanners \cite{Zaccaria2020,Li2019,Mohamed2020,Jia2021,Lecking2006,He2010}. 2D laser rangefinders were used in \cite{Mohamed2020}, where the data was run through a two-stage system consisting of an R-CNN detector and CNN classifier, with a Kalman filter \cite{Li2016} used to localise and track pallets. This method yielded 99.58\% testing accuracy on 340 labelled scans \cite{Mohamed2020}.

\cite{Jia2021} used a time-of-flight camera to collect point cloud information to run through a ResNet \cite{He2016} model. Pallet pocket locations are extracted from the point clouds, and the centres are then determined. The 03D303 IFM Electronics Ltd.\ camera used outputs 93000 distance and grey values for each measurement and is thus highly detailed. The accuracy achieved was up to 94.5\% and included pallets with and without loads, as well as those in stacks or individually positioned.

Pallet detection has also been achieved without machine learning. One such way is by using fiducial markers positioned in the centre and on the sides of the pallets to determine edge positions \cite{Seelinger2005}. Another method makes use of the rectangular shape of pallets for Haar-like features detection \cite{Syu2017}, making use of \textit{OpenCV}\footnote{https://opencv.org/}. Additionally, methods involving the use of ultrasonic chirping, radio waves \cite{Fogel2007}, and RFID tags positioned throughout the environment \cite{Jeon2010}, were used; however, these were found to typically be no more accurate than with sensors just on the trucks \cite{Syu2017}. Machine learning methods offer advantages in the way that no external sensors or marker placements are required to be put on the pallets or any of the warehouse infrastructure besides the camera placed on the vehicle.

\subsection{Synthetic Data}
A general advantage of using synthetic data is anonymity, as there is no personally identifiable information, thus avoiding privacy, ethical and legal issues \cite{Hittmeir2019}. It also removes the need for manual image collection for training data in the ideal case.
Synthetic data has been shown to achieve high accuracy in the right environments, such as 87\% in inventory tracking \cite{Falcao2021}, and about 90\% for CSV data predictions \cite{Hittmeir2019}. In one example, adding labelled synthetic data had practically no effect on the quality of classification \cite{Kuchin2020}.

There are several available synthetic data generation solutions, including those provided commercially by \textit{mostly.ai}\footnote{https://mostly.ai/}, or, the Synthetic Data Vault\cite{Patki2016} or Data Synthesizer\cite{Ping2017}. Data Synthesizer is proven to be suitable for the generation of text-based data sets by\ \cite{Hittmeir2019}. These tools are geared towards the generation of synthetic text data. \textit{POVRay}\footnote{http://www.povray.org/} is a tool that enables the programming of 3D environments to generate synthetic data.

Additionally, Game engines such as \textit{Unity}\footnote{https://unity.com/} and \textit{Unreal Engine}\footnote{https://www.unrealengine.com/} are becoming popular for synthetic data generation due to their ease of use \cite{Falcao2021}.
The only identified downsides of this approach are the domain gap and the need for human labour to generate 3D models \cite{Falcao2021}.
In most situations where much training data is required, these are likely heavily outweighed by the reduced number of hours needed to label images manually.

Some models trained on synthetic data sets lose significant accuracy when run against actual data. For example, a model that achieved greater than 90\% accuracy on synthetic captcha images was reduced to approximately 0\% on real images \cite{Le2017}. This loss of utility was due to humanly in-perceivable miscalibration. Thus, this kind of difference between synthetic and real data is not acceptable, even if some broadness mismatch is fine \cite{Le2017}.

\subsection{Our contribution}
Prior work has proven methodologies to detect pallets, as well as to solve machine vision problems. Additionally, synthetically generated data has been shown to be effective in machine learning. As such, the research gap that this paper hopes to fill is the detection of objects in real images based on synthetically trained neural networks, with only the use of 2D images, rather than 3D implementations.

\section{Methods}
For our training and validation, a pipeline that included predominantly Unity and Detectron2 was used. An outline of the pipeline can be seen below in Figure \ref{fig:Pipeline}.

\begin{figure}[!htbp]
    \centering
    \includegraphics[scale=0.07]{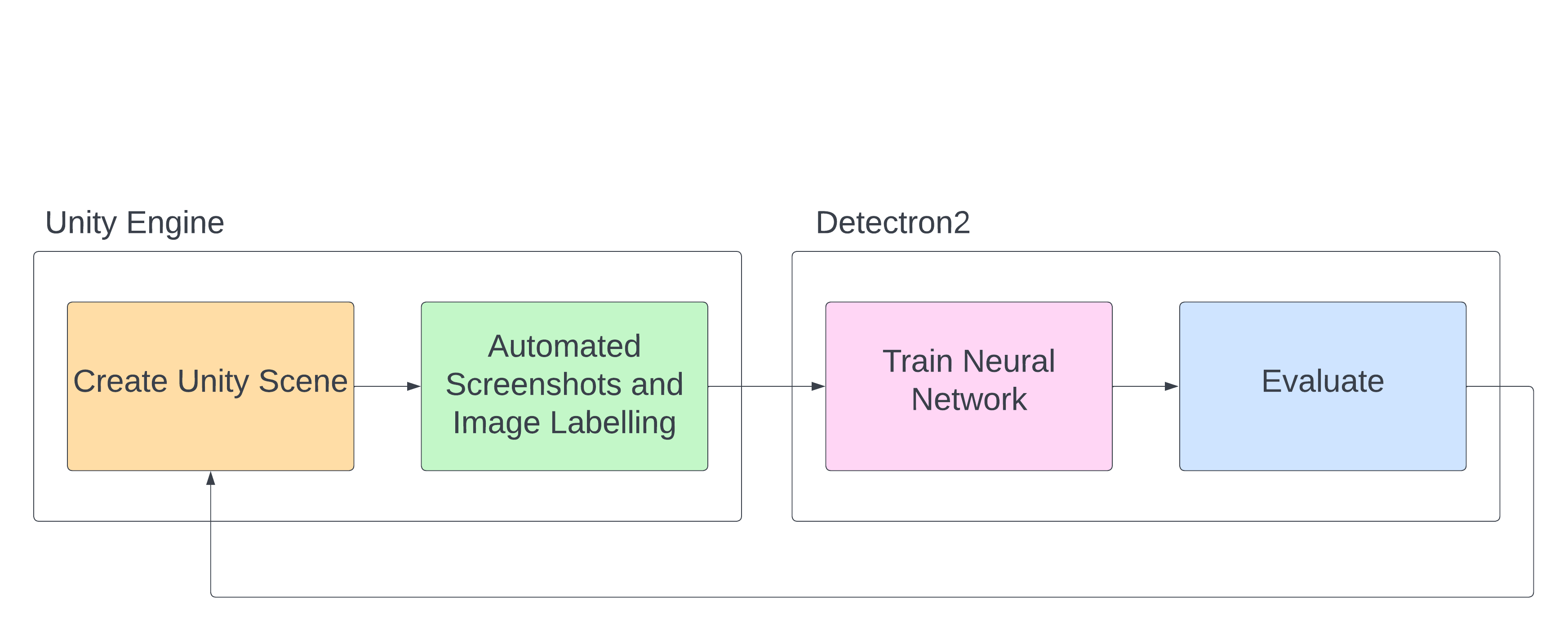}
    \caption{Project pipeline.}
    \label{fig:Pipeline}
\end{figure}

\subsection{Data Generation}
Data generation was handled through the Unity game engine. While normally used for creating games, the engine can be used for other things, such as research and development. Unity was chosen over some of the other 3D rendering software due to its widely available documentation, user support, and longevity in the industry. When compared to other tools like Unreal Engine, which also has some presence in the 3D rendering scene, it was found that unity was easier to use for beginners and to get the initial setup working.

3D models of pallets, racking, and other warehouse paraphernalia are added to a scene and rendered via a 2D camera at certain angles and positions. Two modes were implemented in which to capture these images: manual and automatic. Using the manual mode, the user is able to navigate the camera to any preferred position and angle, whereas in automatic, the camera follows predetermined spherical routes set by the user.

While this process captures the images successfully, a number of algorithms had to be developed to generate COCO-style annotations for each of them.

\begin{figure}[!htbp]
    \centering
    \includegraphics[scale=0.095]{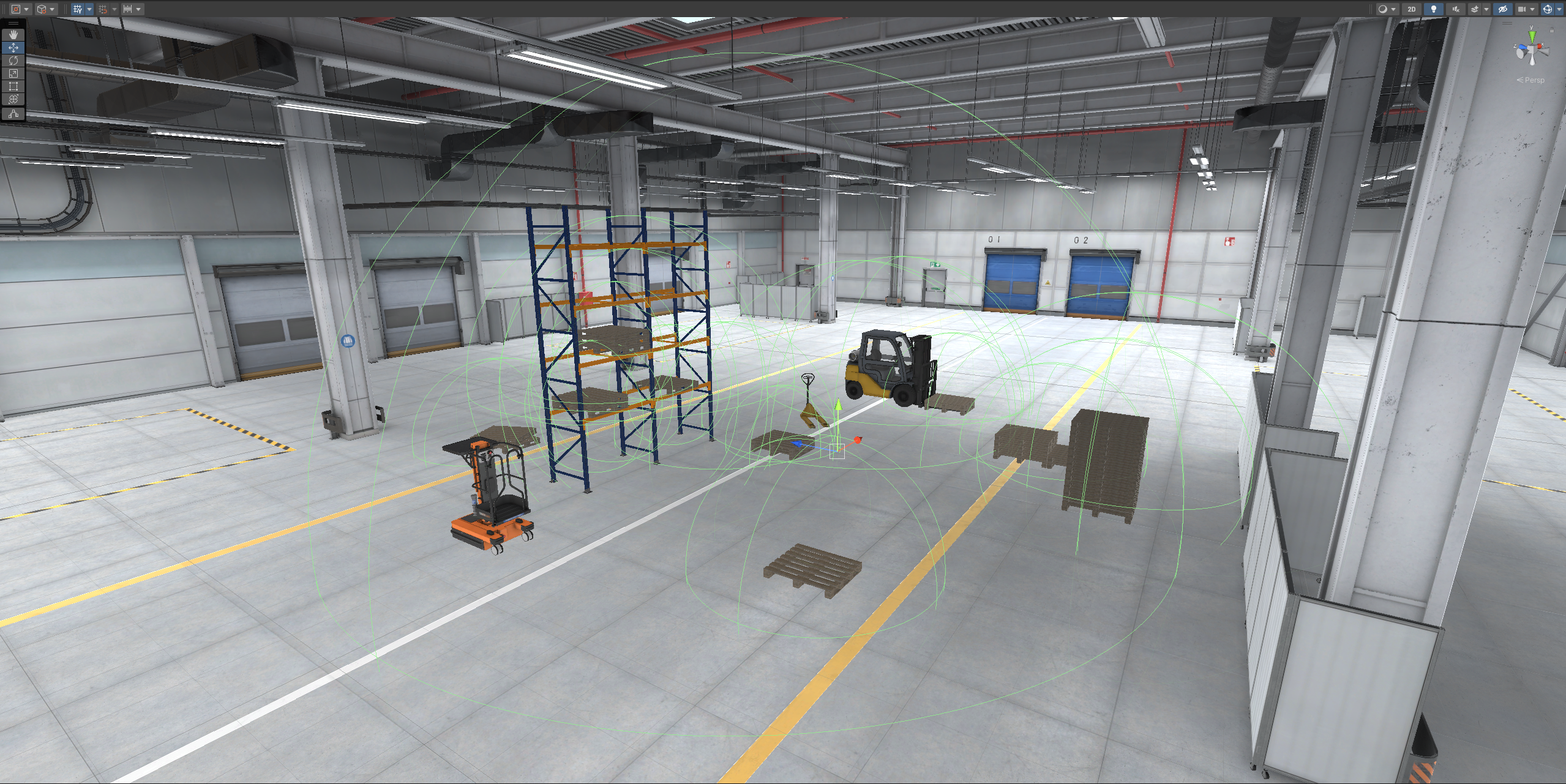}
    \caption{Unity scene used for rendering synthetic data.}
    \label{fig:UnityScene}
\end{figure}

The scene in Unity that was used to render all of the synthetic data can be seen in Figure \ref{fig:UnityScene}, along with all of the "Spheres of Interest" that the camera moved around to capture different angles of the scene for training.

When generating the synthetic data, the generation was broken up into smaller categories based on different scenarios that could occur in a warehouse environment. These included:
\begin{itemize}
    \item Individual Pallets
    \item Stacked Pallets
    \item Pallets on Racking
    \item Pallets on Forklifts
    \item All above combined
\end{itemize}

This was done in order to test if and which configurations of pallets yielded better or worse detection results, or if the configurations of the pallets in the scenes did not matter at all.
It is expected that Individual pallets will be the easiest to detect as these have the most simple features compared to the other scenarios. Stacked pallets and all metrics combined are expected to be the worst as stacked pallets will have less distinguishable features as only the faces are visible and not the body.

\begin{figure}[!htbp]
    \centering
    \includegraphics[scale=0.18]{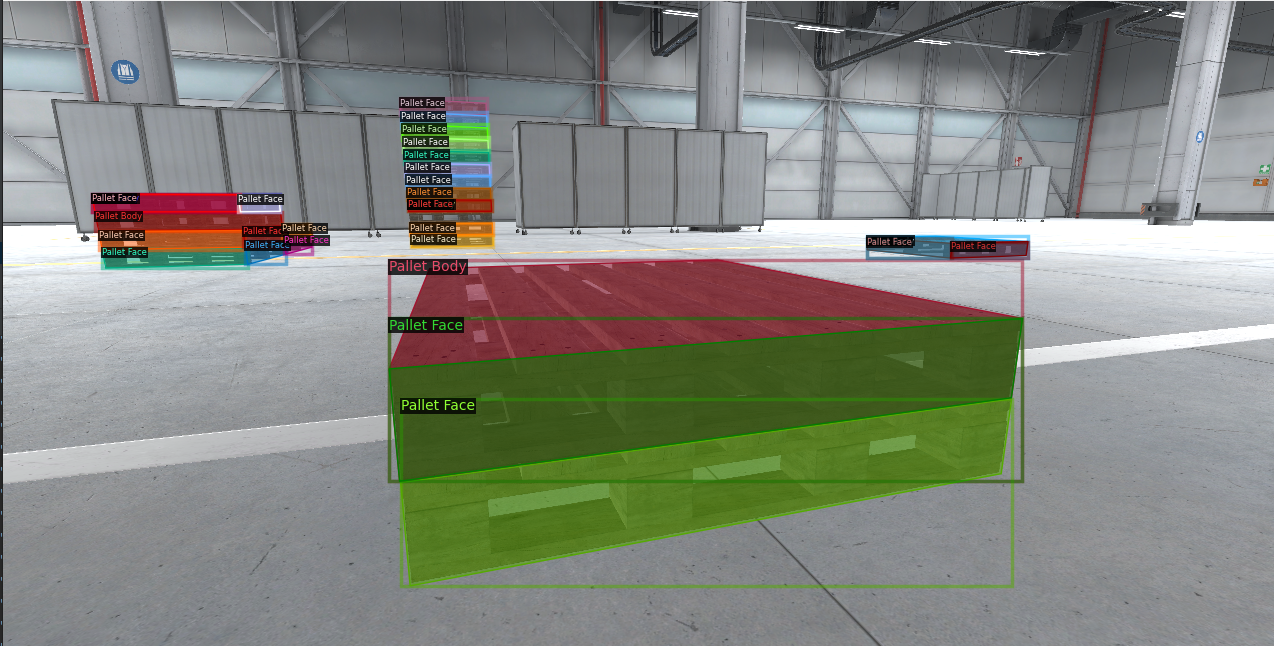}
    \caption{Pallets labelled by Unity engine.}
    \label{fig:UnityLabelling1}
\end{figure}

\begin{figure}[!htbp]
    \centering
    \includegraphics[scale=0.18]{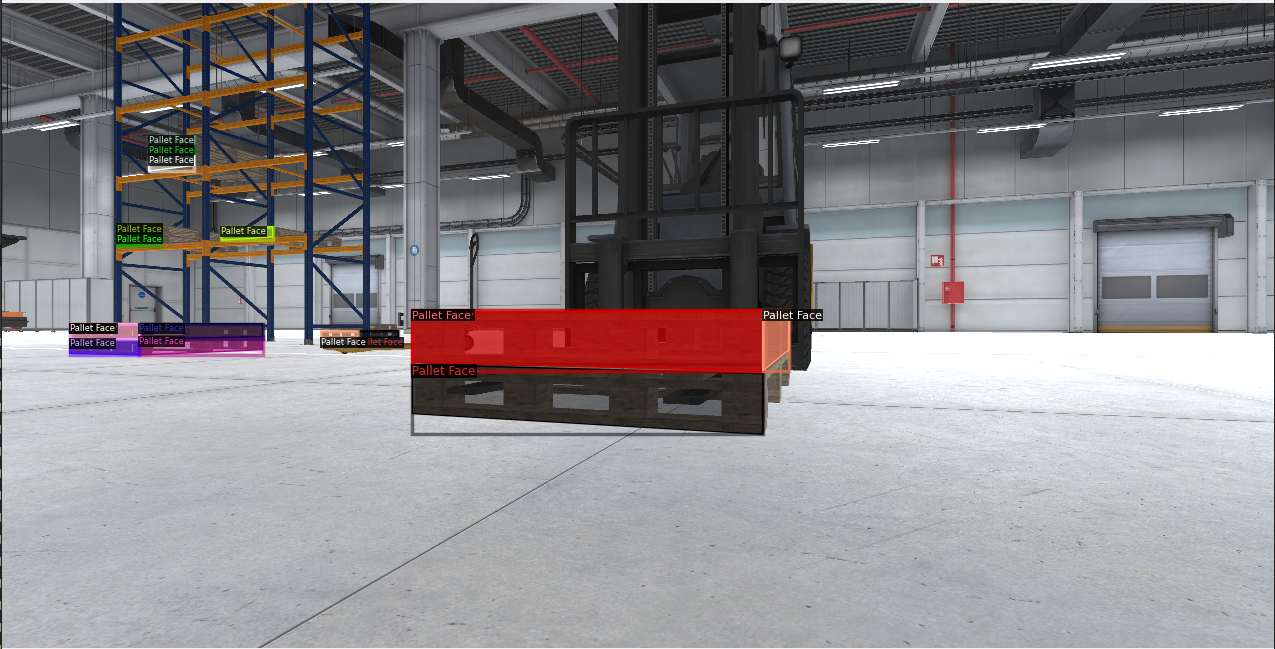}
    \caption{Pallets labelled by Unity engine.}
    \label{fig:UnityLabelling2}
\end{figure}

The end result of these algorithms and synthetic output with unity can be seen in Figures \ref{fig:UnityLabelling1} and \ref{fig:UnityLabelling2} where the synthetic data labels are visualised. While these labels that are present match the images to the exact pixel, some portions of the images are not labelled at all. This is due to limitations in the algorithms that were implemented where pallets that were partially behind objects could not be labelled correctly. This leaves certain parts that a human would have labelled, unlabelled by Unity. An example of a pallet that would not be labelled can be seen below in Figure \ref{fig:UnlabelledPallet}, where the pallet in the back left of the scene would not be labelled since part of it is behind the large stack.

\begin{figure}[!htbp]
    \centering
    \includegraphics[scale=0.12]{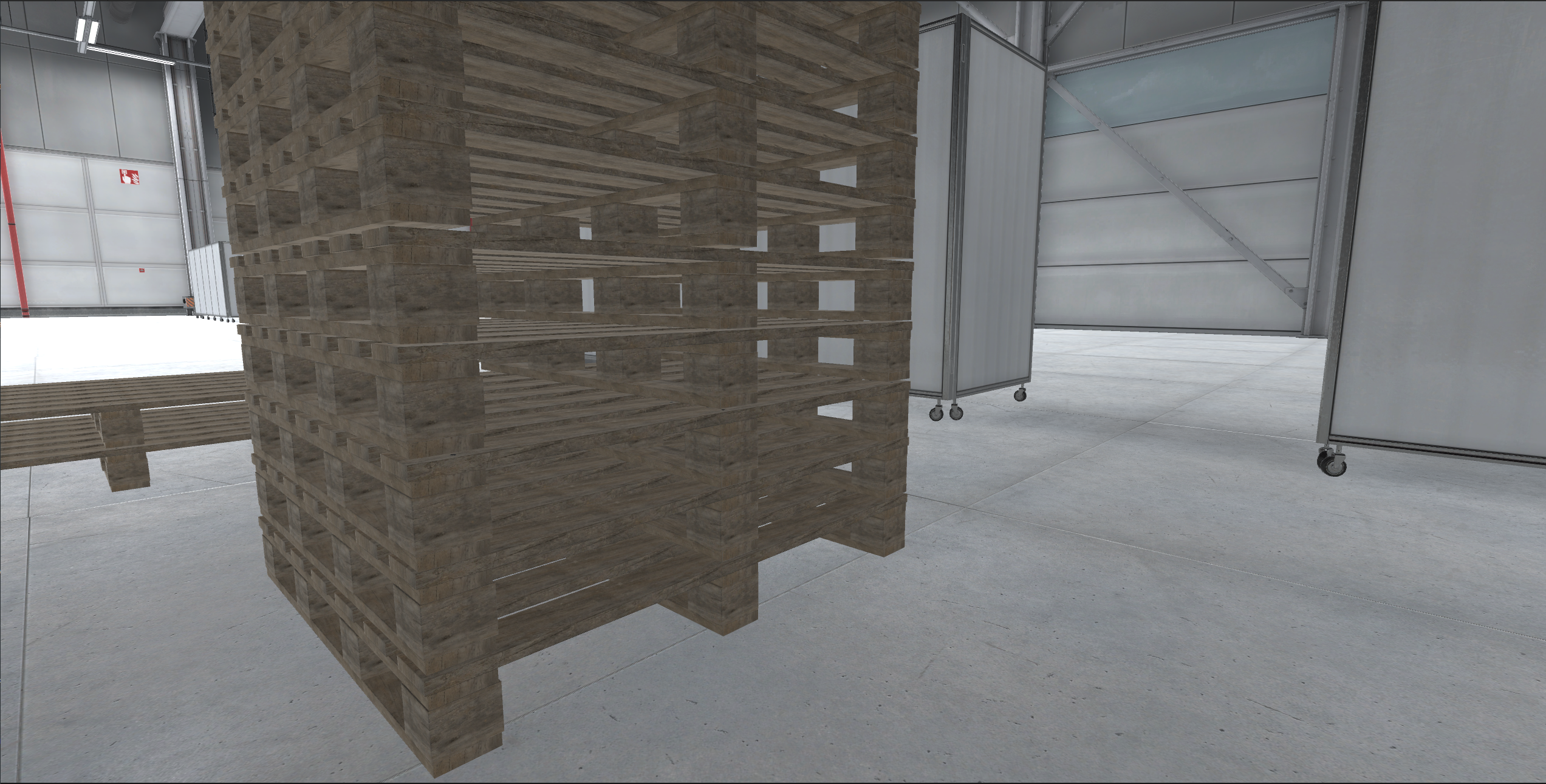}
    \caption{Back left unlabelled pallet.}
    \label{fig:UnlabelledPallet}
\end{figure}

\subsubsection{Pallet Body Detection}

In addition to pallet faces, overall pallet outlines had to be considered. This was particularly challenging as certain vertices had to be added and removed depending on whether the pallet was partially off the screen or if the vertex was not part of the outline. This was done using the Jarvis march algorithm \cite{preparata2012computational}, which can be seen in Figure \ref{fig:ConvexHull} and will be discussed later.

The first step in handling body detection was to remove any pallets which had any vertices that were behind the camera. After this, the 3D points were sorted into a clockwise orientation based on their 2D screen space representation around the centroid of the list of points. This was in preparation for another layer of filtering that needed to be done. Another aspect that needed to be taken into account was the fact that if any points in the outer shape of the pallet were blocked by another pallet, the whole pallet body should be excluded from the labelling. This was because pallet-to-pallet occlusion was to be ignored for the scope of the project. This proved hard to compute as after the vertices had been passed through the Jarvis March algorithm, they were represented as 2D coordinates and, therefore could not be checked if there was anything blocking its view to the camera in 3D space. In order to fix this, the vertices needed to be correlated to their index position and then the whole pallet body label was removed if any of the convex hull elements matched the originals and were blocked from the camera view. The pallet bodies were then run through the screen occlusion algorithm to add or remove additional vertices as required.
\subsubsection{Face Detection}
In order to handle the detection of pallet faces in the synthetic data, several algorithms were developed to determine which vertices needed to be modified in order to handle cases where they could not be seen or where portions of the faces were occluded by the camera. The first step in this process was determining whether the camera had a direct line of sight to the vertices on the face in question. This could either be out of the camera view or within the camera view; as long as no physical object was between the vertex and the camera, the point was included in the filter. After the faces were filtered, each point in the face was checked to see if it was behind the camera; if it was, the whole face was ignored. Each 3D coordinate point was then converted to a 2D screen space coordinate, and an algorithm was run to determine any modifications that needed to be done in order to handle screen occlusion.
\subsubsection{Algorithms Overview}
\begin{figure}[!htbp]
    \centering
    \includegraphics[scale=0.2]{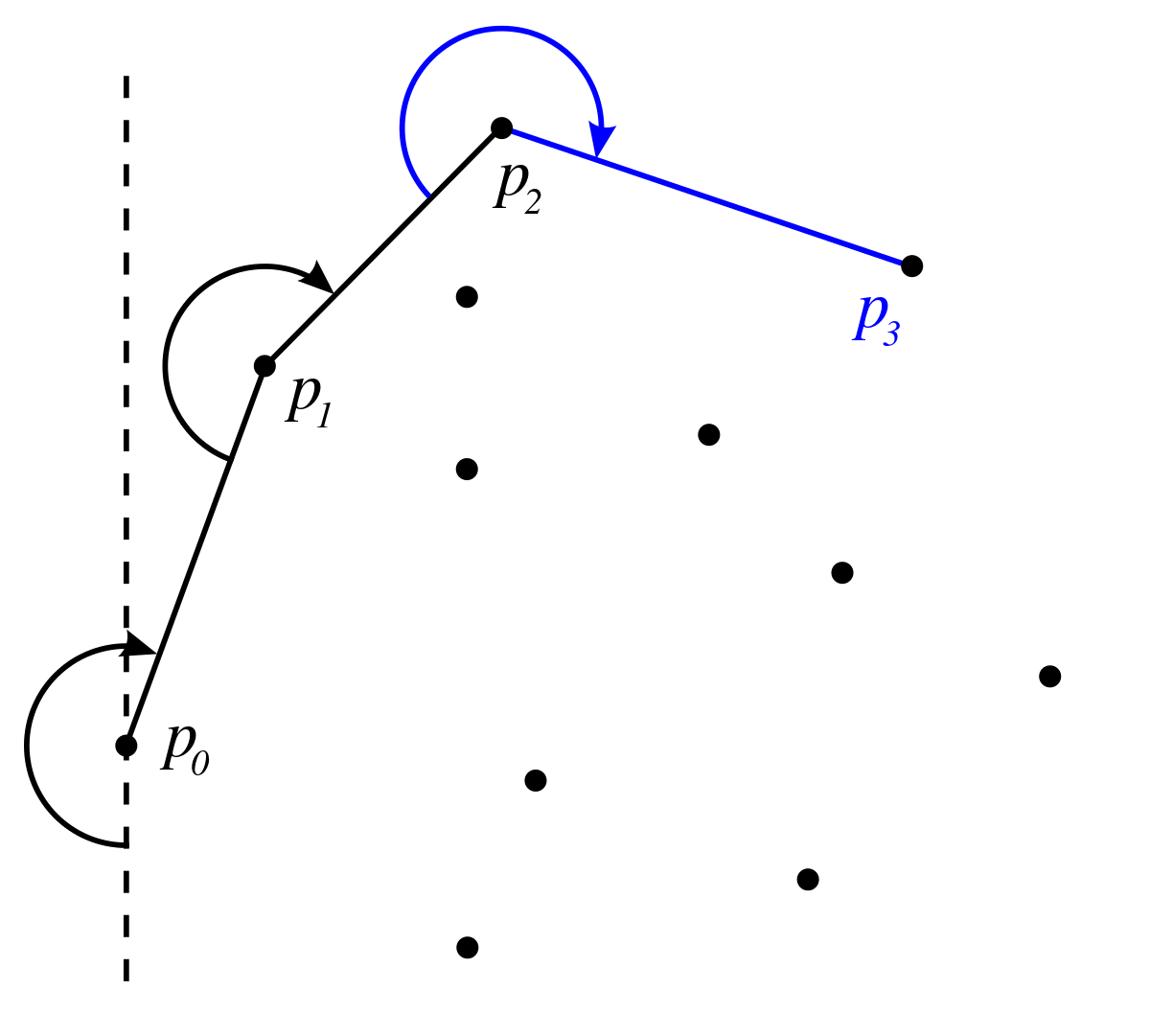}
    \caption{Convex hull visualisation.}
    \label{fig:ConvexHull}
\end{figure}

Several algorithms were used in the creation of the automated labelling data in order to handle the vertex detection within Unity; this was because, while the bounding boxes of each pallet were provided, certain points had to be added or removed if the pallets were partially blocked or off of the screen. The following are some of the algorithms used to handle this task.
\paragraph{Clockwise Point Algorithm}
Clockwise points were handled using the built-in $Atan2$ maths function, which takes an x and y coordinate and returns the angle from the x-axis of the corresponding line from the origin. This angle could then be used to determine the angle of the coordinate point and sort it accordingly. However, this had to be tweaked a little bit to move the $Atan2$ point from the origin to the centroid of all the points. Therefore, the centroid was calculated beforehand and subtracted from all the points to create the required offset.
\paragraph{Jarvis March Algorithm}
The Jarvis March algorithm was used to take a list of coordinates and convert them into a convex hull. Because certain points were not required when labelling the pallet body, these had to be removed. This algorithm worked well for our implementation as the pallet bodies were always convex. An example of excluded points can be seen in figure \ref{fig:ConvexHull}.
\paragraph{Vertex Blocked Check}
Vertices were determined to be blocked or not through the use of ray casts. A single ray is drawn from the camera in the direction of the vertex. If the ray cast collides with anything that is not a vertex on a pallet, the point is determined to be blocked by something in front of the camera.

\subsection{Neural Network}
A base neural network pipeline is required to compare both real-world training data and synthetically generated training data. The chosen pipeline uses Detectron2, with the included COCO Instance Segmentation Mask R-CNN R50-FPN baseline model. This was chosen as it was the most common and easiest to set up to achieve our goal of determining the validity of using synthetic data for machine learning.

Real-world images of pallets were labelled with \textit{Label Studio}\footnote{https://labelstud.io/}, and exported to MS COCO format for use in the aforementioned pipeline. The provided CocoEvaulator calculates quantification metrics, including average precision and average recall.
Images were captured with a high-quality camera in good lighting with an image resolution of 4000x3000 pixels each.

\section{Results}
Multiple sets of synthetic data were generated from automated screen capturing of the Unity scene, as shown in Table \ref{tab:Results}.
The average precision (AP) of pallet faces ($AP_f$) and bodies ($AP_b$) is shown, as well as the overall AP, AP50 and AP75.

Table \ref{tab:Quantities} shows the number of images used for the training and testing of each category.
Additionally, the average recall and F1 score for each category can be seen in Table \ref{tab:Metrics}.

\begin{table}[!h]
    \centering
    \begin{tabular}{|c||c|c|c|c|c|}
        \hline
        &$AP_f$&$AP_b$&AP&AP50&AP75\\ \hline \hline
        Individual&0.45&0.85&0.65&0.86&0.64\\ \hline
        Stacked&0.03&0.04&0.04&0.05&0.04\\ \hline
        On&0.17&0.05&0.11&0.21&0.10 \\
        Racking&&&&& \\ \hline
        On&0.21&0.31&0.26&0.66&0.10 \\ 
        Forklifts&&&&& \\ \hline
        Combined&0.03&0.17&0.10&0.13&0.12 \\ \hline
        Synthetic&0.02&0.18&0.10&0.19&0.10 \\ \hline
    \end{tabular}
    \caption{Classification results (\%).}
    \label{tab:Results}
\end{table}

\begin{table}[!h]
    \centering
    \hspace*{-.5cm}
    \begin{tabular}{|c||c|c|}
        \hline
        & Training & Testing \\ \hline \hline
        Individual & 840 & 15 \\ \hline
        Stacked & 2100 & 7 \\ \hline
        On Racking & 2520 & 37 \\ \hline
        On Forklifts & 1680 & 8 \\ \hline
        Combined & 7140 & 76 \\ \hline
        Combined Synthetic & 4620 & 2520 \\ \hline
    \end{tabular}
    \caption{Data set quantities (number of images).}
    \label{tab:Quantities}
\end{table}

\begin{table}[!h]
    \centering
    \hspace*{-.5cm}
    \begin{tabular}{|c||c|c|}
        \hline
        & Recall & F1 \\ \hline \hline
        Individual & 0.746 & 0.694 \\ \hline
        Stacked & 0.066 & 0.047 \\ \hline
        On Racking & 0.187 & 0.139 \\ \hline
        On Forklifts & 0.386 & 0.314 \\ \hline
        Combined & 0.121 & 0.110 \\ \hline
        Combined Synthetic & 0.184 & 0.130 \\ \hline
    \end{tabular}
    \caption{Average recall and F1 score.}
    \label{tab:Metrics}
\end{table}

\begin{figure}[!htbp]
    \centering
    \includegraphics[scale=0.23]{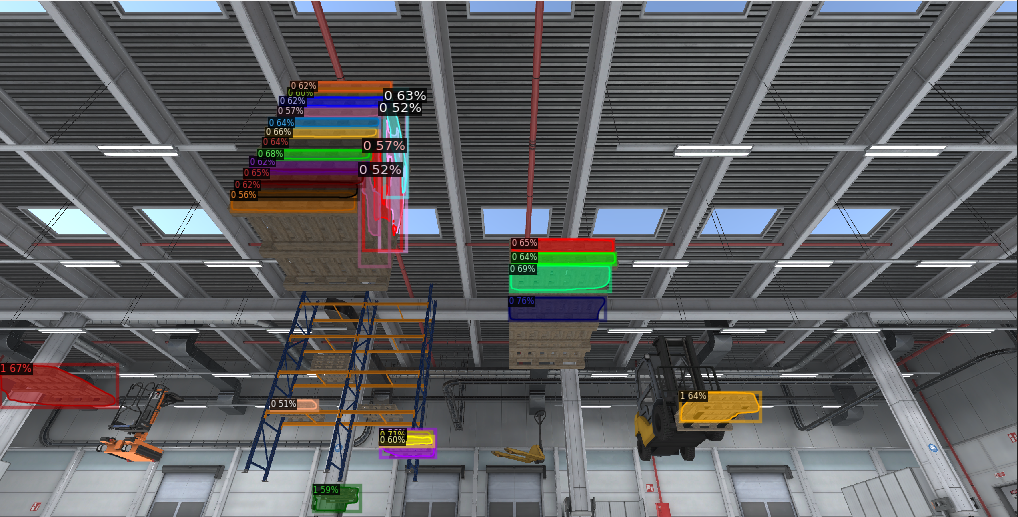}
    \caption{Good synthetic data test result.}
    \label{fig:GoodSyntheticDetection}
\end{figure}

\begin{figure}[!htbp]
    \centering
    \includegraphics[scale=0.23]{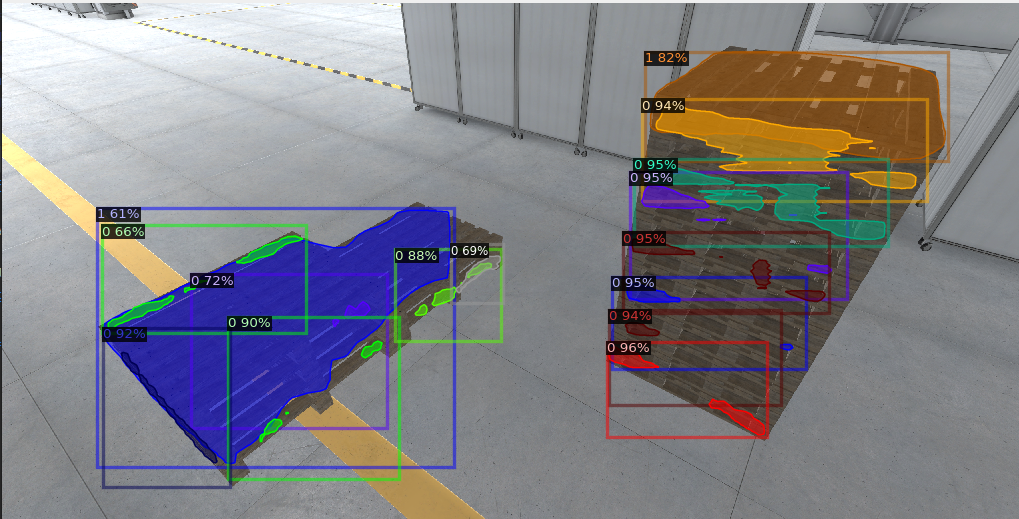}
    \caption{Bad synthetic data test result.}
    \label{fig:BadSyntheticDetection}
\end{figure}

As Table \ref{tab:Results} shows, a model trained on synthetically generated pallets placed by themselves in a warehouse environment, and tested against real-world images of the same, performed the best, yielding an AP50 of 86\%. With an AP50 of 66\%, pallets on forklifts perform well. Stacked pallets and pallets on racking perform poorly in every metric, even AP50. Additionally, test data sets are very small compared to their training data set in certain scenarios. This is due to the limitations of the real images that could be acquired in the given time frame, where only a small portion of them fit the exact categories of interest.
The model, trained on synthetic data, cannot be expected to perform better on real images than on a test set of rendered images. For this reason, in addition to testing the neural network on real images, an overall test was done on a separate synthetic test data set containing all the categories. As can be seen from the results table in \ref{tab:Results}, this did not perform very well, so expectations for the network's performance on real data were not very high. Synthetic detection rates could be improved if they were split into real data categories, as this has seen a noticeable uplift in detection performance. Some of the results of predictions generated by the neural networks can be seen below in Figures \ref{fig:Forklift}, \ref{fig:FaceOnGround}, \ref{fig:GoodSyntheticDetection}, and \ref{fig:BadSyntheticDetection}.

\begin{figure}[!htbp]
    \centering
    \includegraphics[scale=0.15]{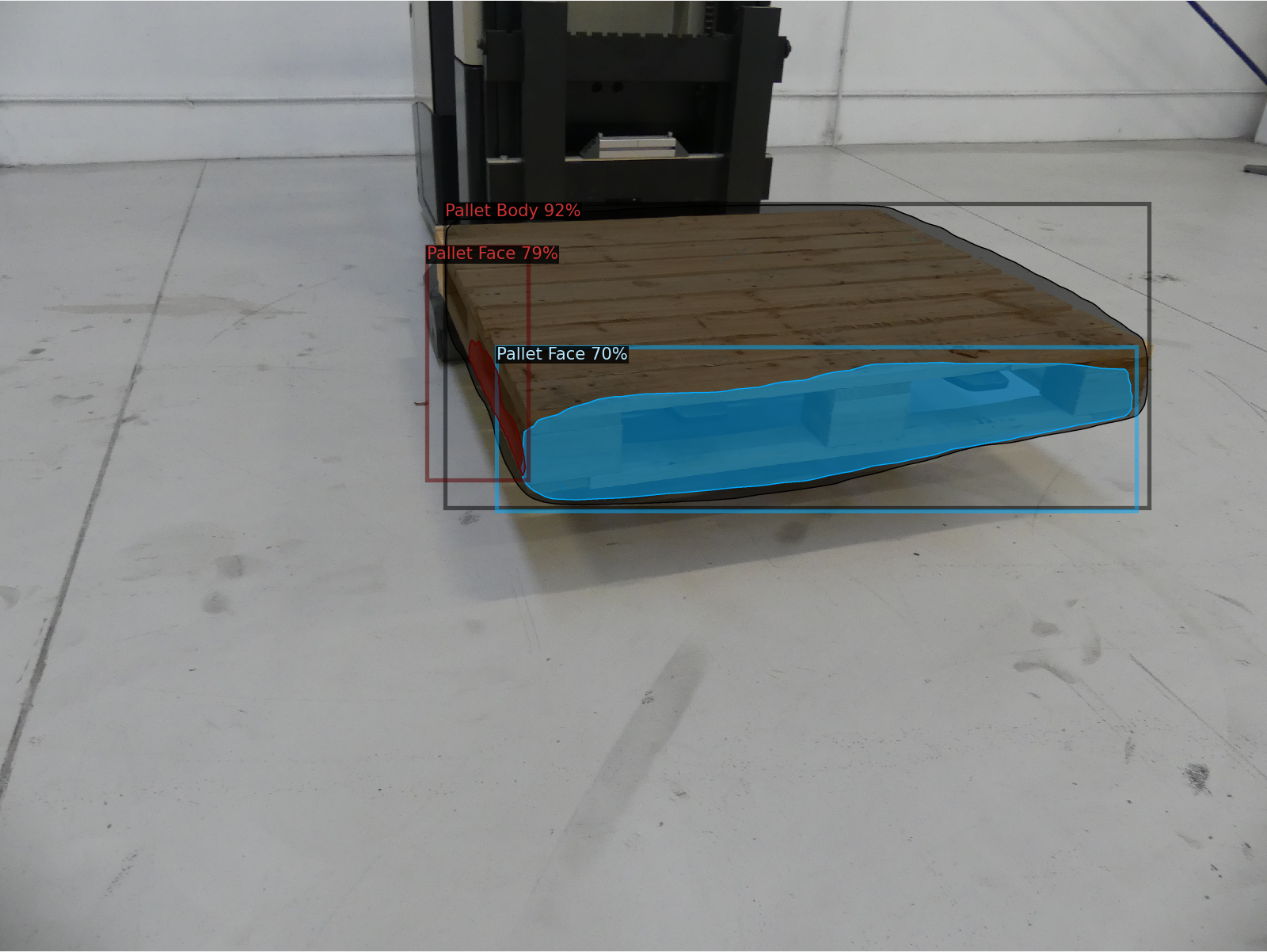}
    \caption{Pallet detection on real forklift.}
    \label{fig:Forklift}
\end{figure}

\begin{figure}[!htbp]
    \centering
    \includegraphics[scale=0.15]{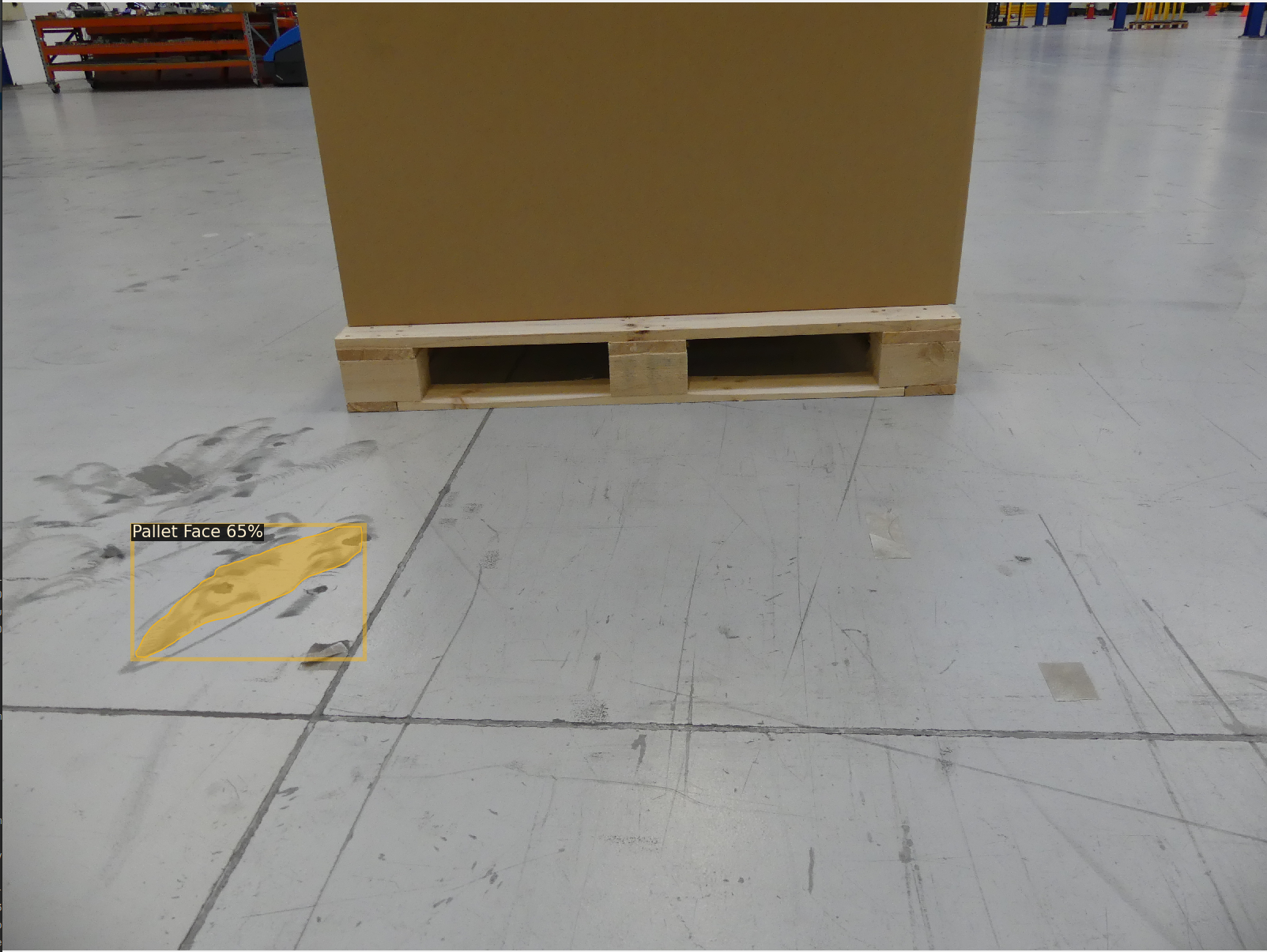}
    \caption{False detection of face on ground.}
    \label{fig:FaceOnGround}
\end{figure}

\section{Discussion}
The selected pallet detection method differs from other methods explored in the literature review. This is due to the fact that other methods used machine learning with other sensors or omitted the use of machine learning and instead used markers or image manipulation.
As seen in Table \ref{tab:Results} and \ref{tab:Metrics}, the results obtained from the neural network varied vastly depending on the scenarios in which they were trained. Individual pallets yield an average precision of 64.9\%, increasing to 85.7\% when looking at the AP50 metric. While these results are good, it can be seen that the precision drops heavily once other scenarios are added into the scenes, such as staked, racking and forklift pallets. The Recall and F1 results in Table \ref{tab:Metrics} confirm this trend, with 75\% and 69\% respectively for individual pallets, dropping substantially for the other scenarios.

One of the major drawbacks identified when using synthetic data for machine learning training was the lack of realism between rendered and real images. This is still a large concern when analysing the results of the neural network here. While the pallets that are used within Unity are very similar to that in the rendering, they are nowhere near identical. This could lead to substantial issues when detecting based on models trained on them. The neural network struggled most with the stacked pallets section. This is thought to be because the features that are available on the label are very similar and are in very close proximity to each other. For example, when the pallets are stacked, only the pallet faces are visible and not the pallet bodies. This results in many pallet face annotations right next to each other, which often confuses the neural network into labelling the entire pallet stack as one pallet body or one pallet face. An example of this can be seen in Figure \ref{fig:UnityLabelling1}.

As mentioned in the results section above, the test data sets for certain categories are very small, while others are substantially larger. This is an additional concern with the accuracy of the trained model, as there may not be enough images in the testing data sets for accurate and meaningful results to be obtained.

Additionally, there is slightly more room to tune the training parameters within Detectron2; this could lead to slightly higher detection rates within the network.

\section{Conclusions and Future Work}
As is made clear by the results obtained, the viability of using synthetically generated data for simple pallet detection is proven.
With an AP50 of 86\% for individual pallets, there is clearly a path forward for industrial use, reducing manual data annotation requirements. However, there is much more work to be done in improving the classification accuracy for scenes in which the pallets are placed in more complex configurations, including being stacked, carrying loads, and obscured behind other objects. Testing data sets could also be improved to provide a more balanced overview of the performance of the neural networks by increasing the number of images inside each category measured.

Due to work being limited by time constraints, it was not possible to write truly comprehensive annotation algorithms for Unity. This functionality can be extended to avoid discarding annotations for a face or body in which any of its edge vertices are blocked by a solid object (pallet, racking, etc.).

There is also the possibility for additional neural network models to be explored rather than just Mask R-CNN, such as the YOLO family, or Swin-L, which performed the best on the MS COCO data set. This could lead to the yield of higher detection rates.

In addition, pallets with loads ought to be accounted for and tested accordingly. Varying models of pallets can also be used, as well as wrapped and otherwise modified pallets. Finally, pinpointing pallet engagement points can be useful for the development of autonomous lifting trucks.

\section*{Acknowledgements}
Thanks to Crown Equipment Corporation for sponsoring this research project and to their representatives Sian Phillips and Abigail Birkin-Hall, for providing the real-world image data set and other invaluable assistance.

Thanks also to Mahla Nejati and Trevor Gee, for supervising the project and providing much-needed insight.

\bibliography{publications}
\bibliographystyle{named}
\end{document}